\documentclass{article}
\usepackage{ijcai25}

\usepackage{times}
\usepackage{soul}
\usepackage{url}
\usepackage[hidelinks]{hyperref}
\usepackage[utf8]{inputenc}
\usepackage[small]{caption}
\usepackage{graphicx}
\usepackage{amsmath}
\usepackage{amsthm}
\usepackage{booktabs}
\usepackage{algorithm}
\usepackage[switch]{lineno}

\usepackage{algpseudocode}
\usepackage{amsfonts}
\usepackage{amsmath}
\usepackage{multicol}

\newtheorem{theorem}{Theorem}

\def\calc{\mathcal{C}}
\def\classes{\calc}
\def\drive{\textsf{drive}}
\def\walk{\textsf{walk}}
\def\bike{\textsf{bike}}
\def\train{\textsf{train}}
\def\bus{\textsf{bus}}

\def\Loss{\mathit{Loss}}
\def\expec{\mathbb{E}}
\def\calm{f_\theta}
\def\conds{C}
\def\prior{\mathbf{\mathcal{P}}}
\def\calo{\mathcal{O}}
\def\calt{\mathcal{T}}
\title{Rule-Based Error Detection and Correction to Operationalize Movement
Trajectory Classification}

\author{
Bowen Xi$^1$
\and
Kevin Scaria$^1$\and
Divyagna Bavikadi$^2$\And
Paulo Shakarian$^2$\\
\affiliations
$^1$Arizona State University\\
$^2$Syracuse University\\
\emails
\{bowenxi, kscaria\}@asu.edu,
\{dbavikad, pshakari\}@syr.edu
}

\begin{document}

\maketitle

\begin{abstract}
Classification of movement trajectories has many applications in transportation and is a key component for large-scale movement trajectory generation and anomaly detection which has key safety applications in environments with unseen movement types.  However, the current state-of-the-art (SOTA) are based on supervised deep learning - which leads to challenges when they encounter novel unseen classes. We provide a neuro-symbolic rule-based framework to conduct error correction and detection of these models to integrate into our movement trajectory platform.  We provide a suite of experiments on several recent SOTA models where we show highly accurate error detection, the ability to improve accuracy on test data that includes novel movement types not seen in training set, and accuracy improvement for the base use case in addition to a suite of theoretical properties that informed algorithm development.  Specifically, we show an F1 scores for predicting errors of up to $0.984$, significant performance increase for unseen movement accuracy ($8.51\%$ improvement over SOTA for zero-shot accuracy), and accuracy improvement over the SOTA model.
\end{abstract}

\section{Introduction}
\label{Introduction}

The identification of a mode of travel for a time-stamped sequence of global position system (GPS) known as ``movement trajectories'' has important applications in travel demand analysis~\cite{huang2019transport}, transport planning~\cite{lin2014mining}, and analysis of sea vessel movement~\cite{fikioris2023optimizing}.  More recently this problem has been of interest for security applications such as leading to efforts such as the IARPA HAYSTAC program\footnote{https://www.iarpa.gov/research-programs/haystac} for which we have created and deployed a platform for trajectory analysis.  A key facet of this problem is the proper classification of trajectories by movement type - particularly in the aftermath of an external shock like a natural disaster.  However, the current current state-of-the-art has relied on supervised neural models~\cite{kim2022gps,dabiri2018inferring} which have been shown to perform well but can experience failure when exposed to previously unseen data, specifically previously unidentified movement types.  In this paper, we extend the current supervised neural methods with a lightweight error detection and correction rule (EDCR) framework providing an overall neurosymbolic system. This framework further enables critical technologies, specifically Artificial Intelligence for Transport, where it's typical to encounter unseen data and require models to not misidentify it. The key intuition is that training and operation data can be used to learn rules that predict and correct errors in the supervised model.  Once trained, the rules are employed operationally in two phases: first detection rules identify potentially misclassified movement trajectories.  A second type of rule to re-classify the trajectories (``correction rules'') is then used to re-assign the sample to a new class.  We present a strong theoretical framework for EDCR rooted in logic and rule mining and formally prove how quantities related to learned rules (e.g., confidence and support) are related to changes in class-level machine learning metrics such as precision and recall. To demonstrate effectiveness empirically, we provide a suite of experiments that show this framework is highly effective in detecting errors (F1 of detecting errors of $0.875$ for the SOTA model, and as high as $0.984$ based on the examined models), unseen movement accuracy of $8.51\%$ over SOTA for zero-shot tuning, and standard classification accuracy improvement over the SOTA model.  In what follows, we provide further background on our domain problem and our current trajectory analysis platform (some of which is a review of \cite{bavikadi2024geospatial}), introduce the algorithmic framework for EDCR including it's theoretical properties, and provide our suite of experimental results before concluding with our findings and future work.

\section{Background}
\begin{figure}[t]
  \centering
  \includegraphics[scale=.23]{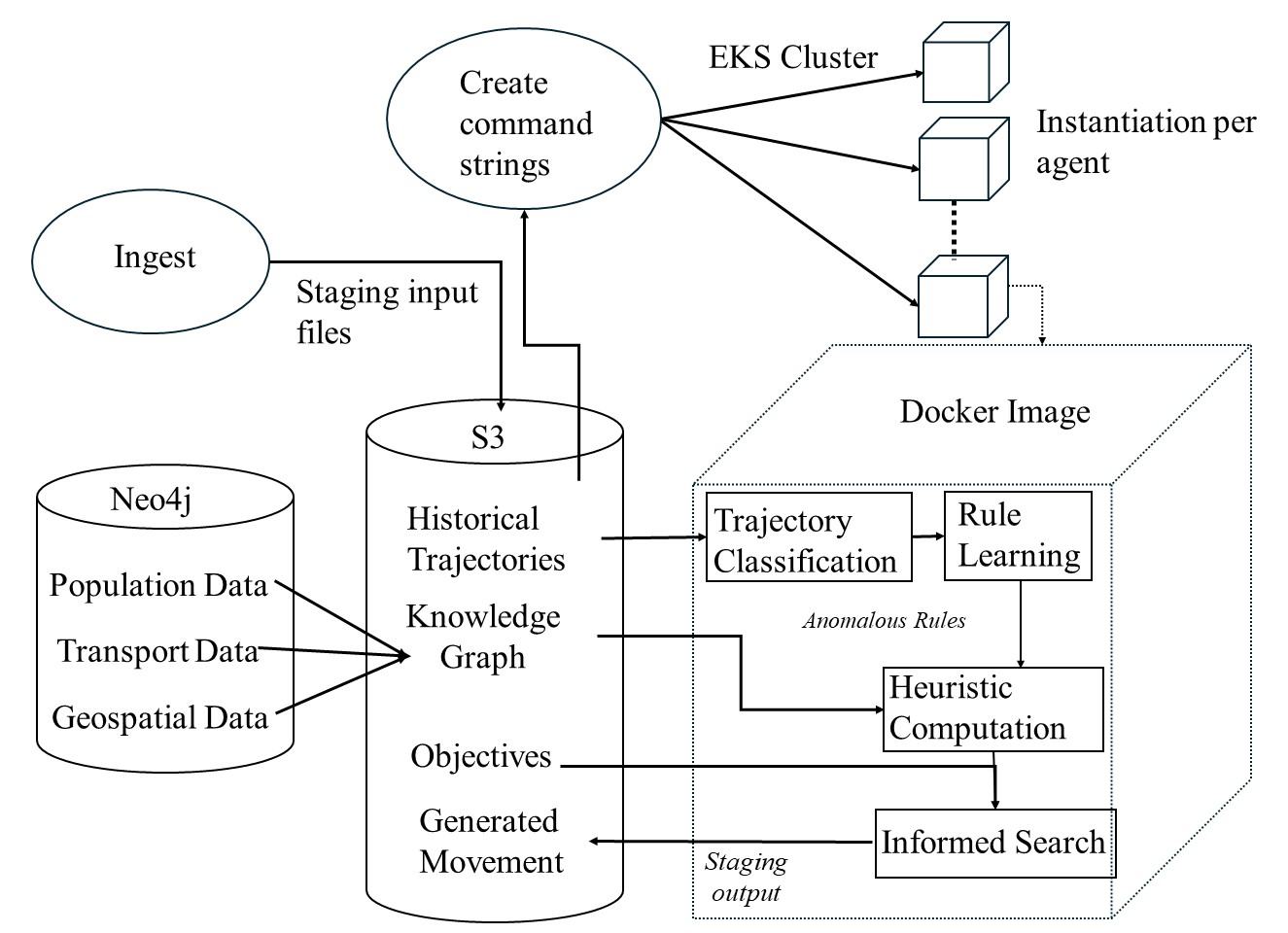}
  \caption{Overall system deployed for government testing}
  \label{fig:dag}
\end{figure}

\noindent\textbf{Overall concept and deployed system.}  Movement types not typically included in the ground truth data emerge with certain target environments (e.g, paid scooters in certain urban areas, auto-rickshaws in South Asia, or boats in Venice). As a result, IARPA (Intelligence Advanced Research Projects Activity) has identified problems relating to the characterization and generation of normal movement as a key problem of study in the HAYSTAC program. Here, the goal is to establish models of normal human movement at a fine-grain level and operationalize those models and techniques in a system deployed to a government environment for evaluation.  As a performer on the program,~\cite{bavikadi2024geospatial} examine the problem of generating realistic movement trajectories.  Initial government tests for trajectory generation involved movement trajectories consisting of only a single mode of transportation.  However, in preparation for the transition to operational use, the government has set requirements to analyze trajectories from various movement types - where the mode of transportation is not known.  As such, we look to operationalize a movement trajectory classification module, which we have depicted in the context of our deployed cloud-based architecture shown in Figure~\ref{fig:dag}.  
This pipeline interfaces with the government system to access the raw geospatial data with related knowledge for various geolocations as well as historical agent trajectories and their corresponding objective files.  Our initial ingest and containerized processes are held in a directed acyclic graph (DAG) as nodes. Our ingest mechanism first parses for the historical trajectories associated with a given agent to stage them in the S3 bucket. Then, geospatial data stored in Neo4j is consolidated into a knowledge graph and staged into the S3 bucket.
We instantiate pods on the Amazon Elastic Kubernetes Service (EKS) cluster for all agents with a Docker image to analyze the staging folders and create the respective string commands specific to each agent. The trajectory classification module identifies and tags the modes of transportation in the corresponding trajectory, which is further used to learn rules while considering different types of movements. These rules along with the knowledge graph are used to compute the heuristic value for an informed search method (A* search) to generate movement trajectories~\cite{bavikadi2024geospatial}. 
As the container runs, generated movement instruction files are pushed to the appropriate output directory as seen in Figure~\ref{fig:dag}. Additionally, the generated movement abides by predefined spatiotemporal constraints (objectives).
\vspace{3pt}

\noindent\textbf{Movement Trajectory Classification Problem.}  The problem of classifying movement trajectories has been studied in the literature~\cite{zheng2008understanding,wang2017detecting,simoncini2018vehicle,dabiri2018inferring,kim2022gps} and we shall refer to it as the movement trajectory classification problem (MTCP).  We also note that this line of work differs from and is complementary to trajectory generation~\cite{bavikadi2024geospatial,janner2021sequence,chen21,itkina2022interpretable} which does not seek to identify the mode of transportation.  An MTCP instance is defined as given a sequence of GPS points, $\omega$, assign one of $n$ movement class from $\calc$ which is often defined~\cite{kim2022gps,dabiri2018inferring} as $\calc=\{\walk, \bike, \bus, \drive, \train\}$.  The current paradigm for the MTCP problem is to create a neural model $f_\theta$ that maps sequences to movement classes using a set of weights, $\theta$.  In this approach traditional methods (i.e., gradient descent) are used to find a set of parameters such that a loss function is minimized based on some training set $\calt$ (where each sample $\omega \in \calt$ is associated with a ground truth class $gt(\omega)$).  Formally: $\arg \min_\theta \expec_{\omega \in \calt}\Loss(f_\theta(\omega),gt(\omega))$.  Within this paradigm, several approaches have been proposed.  Most notably a CNN-based architecture~\cite{dabiri2018inferring} and the current state-of-the-art approach known as Long-term Recurrent Convolutional Network (LRCN)~\cite{kim2022gps} which combines lower CNN layers and upper LSTM layers - both of which we use as baselines in addition to an extension of LRCN that uses an additional attention head (LRCNa).  
\vspace{3pt}

\noindent\textbf{Limitations of Current SOTA.}  However, there are several limitations to these approaches that are problematic in the context of the IARPA HAYSTAC use case.
\begin{itemize}
    \item \textit{Not designed for unseen movements.} Any supervised MTCP model requires a data set whose movement classes match the target environment.  To address the more dynamic needs of our government customer, we require approaches that can identify when they are likely to give incorrect results to adapt to novel environments.
    \item \textit{All classes known a-priori.} In the prior work, set $\calc$ is treated as static and complete meaning that novel movement types not in training set will not be properly classified and not identified as being different than a movement type in class $\calc$. 
    \item \textit{Previous Results Evaluated on Overlapping Training and Testing Sets.}  As noted in \cite{zeng_trajectory-as--sequence_2023}, the standard evaluation of MTCP approaches has been on datasets that experience leakage between train and test.  Due to the operational nature of this work, we must examine other splits.
\end{itemize}
Deploying movement trajectory classification models to a certain environments can lead to movements not seen in training (e.g., paid scooters are not seen in training but prevalent in certain urban areas).  Hence, these ``novel movements" will inherently be classified incorrectly.
The common element in all of these limitations is an understanding of when such classifiers are likely wrong. 
However, this goes beyond retraining or selecting from different training data as the government customer envisions use-cases with unseen movement types- hence training data would be limited. This generally precludes meta-learning and domain generalization~\cite{hospedales2021meta,zhou2022domain,vanschoren2018meta,maes1988meta} which attempt to account for changes in the distribution of data and/or selection of a model that was trained on data similar to the current problem.  Additionally, these problems must also be addressed in the context of our existing system (Figure~\ref{fig:dag}), which employs symbolic reasoning to generate movement trajectories - ensuring they attain a degree of normalcy~\cite{bavikadi2024geospatial}.  As a result, we examined approaches for characterizing failures in machine learning models such as introspection~\cite{daftry_introspective_2016,ramanagopal_failing_2018}, however, these approaches only predict model failure and do not attempt to explain or correct it.  Another area of related work is machine learning verification that~\cite{ivanov_verisig_2021,jothimurugan_compositional_2021,ma2020stlnet}) that looks to ensure the output of an ML model meets a logical specification - however to-date this work has not been applied to correct the output of a machine learning model and generally depend on the logical specifications being known a-priori (not an assumption we could make for our use case).  In recent studies on abductive learning \cite{huang2023enabling,dai2019bridging} and neural symbolic reasoning \cite{cornelio2022learning}, incorporate error correction mechanisms rooted in inconsistency with domain knowledge as logical rules - but as with verification, we do not have this symbolic knowledge a-priori.

\section{Error Detection and Correction Rules}
\label{sec:errorCorr}

To address the issues of the previous section, we are employing a rule-based approach to correcting MTCP model $f_\theta$.  The intuition is that using limited data, we will learn a set of rules (denoted $\Pi$) that will be able to detect and correct errors of $f_\theta$ by logical reasoning~\cite{aditya23}.  Then, upon deployment for some new sequence $\omega$, we would first compute the class $f_\theta(\omega)$ and then use the rules in set $\Pi$ to conclude if the result of $f_\theta$ should be accepted and if not, provide an alternate class in an attempt to correct the mistake.  In this section, we formalize the error correcting framework with a simple first order logic (FOL) and provide analytical results relating aspects of learned rules that inform our analytical approach to learning such error detecting and correcting rules. We complete the section with a discussion on how various potential ``failure conditions'' are extracted to create the rules to correct errors.

In this paper, we shall assume a set $\calo$ of operational sequences for which there is ground truth available after model training.  This set can be the set of training data, a subset, or a superset.  Later, in our experiments, we look at cases where $\calo=\calt$ and $\calt \subseteq \calo$ - however these are not requirements as our results are based on model performance on $\calo$ - and we envision use-cases where $\calo$ is significantly different from $\calt$. On these samples, for each class $i$, the model ($f_\theta$) returns class $i$ for $N_i$ of the samples, and for each class $i$ we have the number of true positives $TP_i$, false positives $FP_i$, true negatives $TN_i$, and false negatives $FN_i$.  We have precision $P_i=TP_i/N_i = TP_i/(TP_i + FP_i)$, recall $R_i=TP_i/(TP_i+FN_i)$, and prior of predicting class $i$: $\prior_i = N_i/N$.

\noindent\textbf{Language.}  We use a simple first-order language where samples are represented by constant symbols ($\omega$).  We define set $\conds$ of $m$ ``condition'' unary predicates $cond_1,\ldots,cond_m$ associated with each sample that can be either true or false - these are conditions that can be thought of as potentially leading to failure (but our learning algorithms will identify which ones lead to failure for a given prediction).  We also define unary predicates for each class $i$: $pred_i$, $corr_i$, and $error$ defined below.
\begin{itemize}
\item $pred_i$: True if and only if the model predicts class $i$ i.e., $pred_i(\omega)$ is true iff $f_\theta(\omega)=i$.
\item $corr_i$: This predicate is true if and only if the correct movement class for $\omega$ is $i$, i.e., $corr_i(\omega)$ is true iff $gt(\omega)=i$.
\item $error$: This predicate is true if and only if an EDCR rule concludes there is an error in the model's prediction.
\end{itemize}

\noindent\textbf{Rules.}  The set of rules $\Pi$ will consist of two rules for each class: one ``error detecting'' and one ``error correcting.''  Error detecting rules which will determine if a prediction by $\calm$ is not valid.   In essence, we can think of such a rule as changing the movement class assigned by $\calm$ to some sample $\omega$ from $i$ to ``unknown.''  For a given class $i$, we will have an associated set of detection conditions $DC_i$ that is a subset of conditions, the disjunction of which is used to determine if $\calm$ gave an incorrect classification. 
\begin{eqnarray}
error(\omega) \leftarrow pred_i(\omega) \wedge \bigvee_{j\in DC_i}cond_j(\omega)
\end{eqnarray}

After the application of the error detection rules for each class, we may consider re-assigning the samples to another class using a second type of rule called the ``corrective rule.''  Such rules are formed based on a subset of conditions-class pairs $CC_i \subseteq \conds \times \classes$. 
\begin{eqnarray}
corr_i(\omega) \leftarrow \bigvee_{q,r \in CC_i}\left( cond_q(\omega) \wedge pred_r(\omega) \right)
\end{eqnarray}
Associated with the rules of both types are the following values - both are defined as zero if there are no conditions.
\vspace{2pt}

\noindent\textit{Support ($s$):} fraction of samples in $\calo$ where the body is true.\vspace{2pt}

\noindent\textit{Support w.r.t. class $i$ ($s_i$):} given the subset of samples where the model predicts class $i$, the fraction of those samples where the body is true (note the denominator is $N_i$).\vspace{2pt}

\noindent\textit{Confidence ($c$):} the number of times the body and head are true together divided by the number of times the body is true.\vspace{2pt}

Now we present some analytical results that inform our learning algorithms.  Our strategy for learning involves first learning detection rules (which establish conditions for which a given classification decision by $\calm$ is deemed incorrect) and then learning correction rules (which then correct the detected errors by assigning a new movement class to the sample).  We formalize these two tasks as follows.
\vspace{2pt}

\noindent\textit{Improvement by error detecting rule.} For a given class $i$, find a set of conditions $DC_i$ such that precision is maximized and recall decreases by, at most $\epsilon$.
\vspace{2pt}

\noindent\textit{Improvement by error correcting rule.} For a given class $i$, find a subset $CC_i$ of $\conds \times \classes$ such that both precision and recall are maximized.
\vspace{2pt}

\noindent\textbf{Properties of Detection Rules.}  First, we examine the effect on precision and recall when an error detecting rule is used.  Our first result shows a bound on precision improvement.  If class support ($s_i$) is less than $1-P_i$, which we would expect (as the rule would be designed to detect the $1-P_i$ portion of results that failed), then we can also show that the quantity $c\cdot s_i$ gives us an upper bound on the improvement in precision.\footnote{Complete proofs for all formal results can be found at \\ \url{https://arxiv.org/abs/2308.14250}.}

\begin{theorem}
\label{thm:precNeg}
Under the condition $s_i \leq 1-P_i$, the precision of model $\calm$ for class $i$, with initial precision $P_i$, after applying an error detecting rule with support $s_i$ and confidence $c$ increases by a function of $s_i$ and $c$ and is no greater than $c\cdot s_i$ and this quantity a normalized polymatroid submodular function with respect to the set of conditions in the rule $DC_i$.
\end{theorem}

The error detecting rules can cause the recall to stay the same or decrease.  Our next result tells us precisely how much recall will decrease.

\begin{theorem} 
\label{thm:rec_thm}
After applying the rule to detect errors, the recall will decrease by $(1-c)s_i\frac{R_i}{P_i}$ and this quantity is a normalized polymatroid submodular function with respect to the set of conditions in the rule $DC_i$.
\end{theorem}

\begin{algorithm}[h]
\caption{\textsf{DetRuleLearn}}
\label{alg:ruleSelct}
\begin{small}
\begin{algorithmic}
    \State {\bfseries Require:} Class $i$, Recall reduction threshold $\epsilon$, Condition set $\conds$
    \State {\bfseries Ensure:} Subset of conditions $DC_i$
    \State{$DC_i:=\emptyset$}
    \State{$DC^* := \{ c \in \conds \textit{ s.t. } NEG_{\{c\}} \leq \epsilon \cdot \frac{N_iP_i}{R_i} \}$ }
    \While{$DC^* \neq \emptyset$}
        \State{$c_{best}=\arg\max_{c \in DC^*} POS_{DC_i\cup\{c\}}$}
        \State{Add $c_{best}$ to $DC_i$}
        \State{$DC^* :=\{ c \in \conds\setminus DC_i\textit{ s.t. }  NEG_{DC_i\cup\{c\}} \leq \epsilon \cdot \frac{N_iP_i}{R_i} \}$}
    \EndWhile
    \State{\textbf{return} $DC_i$}
\end{algorithmic}
\end{small}
\end{algorithm}

As the quantities identified Theorems~\ref{thm:precNeg} and \ref{thm:rec_thm}  are submodular and monotonic, we can see that the selection of a set of rules to maximize $c\cdot s_i$ subject to the constraint that $(1-c)s_i\frac{R_i}{P_i}\leq \epsilon$ is a special case of the ``Submodular Cost Submodular Knapsack'' (SCSK) problem and can be approximated with a simple greedy algorithm~\cite{iyer13} with approximation guarantee of polynomial run time (Theorem 4.7 of \cite{iyer13}).  Our algorithm \textsf{DetRuleLearn} is an instantiation of such an approach to creating an error detecting rule for a given class that maximize precision while not reducing recall more than $\epsilon$. Here, $\epsilon$ is treated as a hyperparameter.  Also, $POS_{DC}$ and $NEG_{DC}$ are simply the number of samples that satisfy the conditions for some set $DC$ and are true errors (for $POS_{DC}$) and non-errors (for $NEG_{DC}$).  In other words, given a set of condition class pairs and the rule of interest, $BOD$ here is the number of examples that satisfy the body (class-condition pair) of the error detection rules, and $POS$ here is the number of examples that satisfy the body (class-condition pair) and the head of the error detection rules. $P_i, R_i$ are precision and recall for class $i$ while $N_i$ is the number of samples that the model classifies as class $i$.

\noindent\textbf{Properties of Corrective Rules.} In what follows, we shall examine the results for corrective rules.  Here, the error correcting rule with predicate $corr_j$ in the head will have a disjunction of elements of set $CC_i \subseteq \conds \times \classes$.  Also, note that here the support $s$ is used instead of class support ($s_i$).  Here we find that both precision and recall increase with rule confidence (Theorem~\ref{thm:posThm}).

\begin{theorem}
\label{thm:posThm}
For the application of error correcting rules, both precision and recall increase if and only if rule confidence ($c$) increases.
\end{theorem}

This result suggests that optimizing confidence will optimize both precision and recall.  However, this is not a monotonic function over $CC_i$, so we adopt a fast, heuristic approach for non-monotonic optimization based on \cite{buchbinder12}, presented by \textsf{CorrRuleLearn} in this paper.  Here, we will consider an initial set of condition-class pairs $CC_{all}$ that is a subset of $\conds \times \classes$.  For a given class for which we create an error correcting rule, we select $CC_i$ from this larger set using our approach.  Note here that $POS_{CC}$ is the number of samples that satisfy the rule body and head ($corr_i(\omega)$ in this case) given a set of condition-class pairs $CC$ while $BOD_{CC}$ is the number of samples that satisfy the body formed with set $CC$.
\vspace{5pt}

\begin{algorithm}[h]
\caption{\textsf{CorrRuleLearn}
\label{alg:posRuleSelct}}
\begin{small}
\begin{algorithmic}
\State {\bfseries Require:} Class $i$, Set of condition-class pairs $CC_{all}$
\State {\bfseries Ensure:} Subset of condition-class pairs $CC_i$
\State{$CC_i := \emptyset$}
\State{$CC_i' := CC_{all}$}
\State{Sort each $(c,j)\in CC_{all}$ from greatest to least by $\frac{POS_{\{(c,j)\}}}{BOD_{\{(c,j)\}}}$ and remove $\frac{POS_{\{(c,j)\}}}{BOD_{\{(c,j)\}}} \leq P_i $ }

\For{$(c,j) \in CC_{all}$ selected in order of the sorted list}
    \State $a := \frac{POS_{CC_i \cup \{(c,j)\}}}{BOD_{CC_i\cup \{(c,j)\}}}-\frac{POS_{CC_i}}{BOD_{CC_i}}$
    \State $b := \frac{POS_{CC_i' \setminus \{(c,j)\}}}{BOD_{CC_i' \setminus \{(c,j)\}}}-\frac{POS_{CC_i'}}{BOD_{CC_i'}}$
    \If{$a \geq b$}
        \State{$CC_i := CC_i \cup \{(c,j)\}$}
    \Else
        \State{$CC_i':= CC_i' \setminus \{(c,j)\}$}
    \EndIf
\EndFor

\If{$\frac{POS_{CC_i}}{BOD_{CC_i}}\leq P_i$}
    \State{$CC_i := \emptyset$}
\EndIf
\State \textbf{return} $CC_i$
\end{algorithmic}
\end{small}
\end{algorithm}

\begin{algorithm}[h]
\caption{\textsf{DetCorrRuleLearn}\label{alg:ruleLearn}}
\begin{algorithmic}
\State {\bfseries Require:} Recall reduction threshold $\epsilon$, Condition set $\conds$
\State {\bfseries Ensure:} Set of rules $\Pi$
\State $\Pi := \emptyset$
\State $CC_{all} := \emptyset$
\For{Each class $i$}
    \State{$DC_i := \textsf{DetRuleLearn}(i, \epsilon, \conds)$}
    \If{$DC_i \neq \emptyset$}
        \State{$\Pi := \Pi \cup$}
        \State{\,$\{ error(\omega) \leftarrow pred_i(\omega) \wedge \bigvee_{j\in DC_i}cond_j(\omega)\}$}
    \EndIf
    \For{$cond \in DC_i$}
        \State{$CC_{all}:=CC_{all}\cup \{(cond, i) \}$}
    \EndFor
\EndFor
\For{Each class $i$}
    \State{$CC_i := \textsf{CorrRuleLearn}(i, CC_{all})$}
    \If{$CC_i \neq \emptyset$}
        \State{$\Pi := \Pi \cup$}
        \State{\,$\{ corr_i(\omega) \leftarrow \bigvee_{q,r \in CC_i}\left( cond_q(\omega) \wedge pred_r(\omega) \right)\}$}
    \EndIf
\EndFor
\State \textbf{return} $\Pi$
\end{algorithmic}
\end{algorithm}

\noindent\textbf{Learning Detection and Correction Rules Together.}  Error correcting rules created using \textsf{CorrRuleLearn} will provide optimal improvement to precision and recall for the rule in the target class, but in the case of multi-class problems, it will cause recall to drop for some other classes.  However, we can combine error detecting and correcting rules to overcome this difficulty.  The intuition is first to create error detecting rules for each class, which effectively re-assigns any sample into an ``unknown'' class.  Then, we create a set $CC_{all}$ (used as input for \textsf{CorrRuleLearn}) based on the conditions selected by the error detecting rules.  In this way, we will not decrease recall beyond what occurs in the application of error detecting rules.

\noindent\textbf{Algorithmic Efficiency.} We note that these algorithms are quite efficient.  For example, \textsf{DetRuleLearn} is quadratic in the number of conditions and linear in the number of samples.  However, in practice it actually performs better, as the outer loop iterates significantly less than the total number of conditions and the number of selected conditions is reduced with each iteration.  Likewise, the algorithm \textsf{CorRuleLearn} is linear in the number of samples and linear in the number of condition-class pairs.    

\noindent\textbf{Conditions for Error Detection and Correction.}  Practically, the source of the conditions from which our algorithms create EDCR rules (set $C$) needs to be instantiated.  We adopt two straightforward approaches to this.  First, we use a binary version of the classifier -- for given class $i$, we have a binary classifier $g_i$ which returns ``true'' for sample $\omega$ if $g_i$ assigns it as $i$ and ``false'' otherwise. In this way, for each sample $\omega$ we have a $g_i(\omega)$ condition for each of the classes.  The second way we create conditions is based on outlier behavior based on the velocity of the vehicle in the sample.  Here, if the velocity of a given sample is above a threshold (based on the maximum value for ground truth in the training data) this velocity condition is true - and it is false otherwise.

\section{Experimental Evaluation}
\label{sec:expSec}

\noindent\textbf{Experimental Setup.}  Previous work such as \cite{kim2022gps} is known to have data leakage based on the split between training and test primarily due to segments of a movement sequence existing in both training and test sets~\cite{zeng_trajectory-as--sequence_2023}.  In this paper, we examine a training-test split with no overlap between the two avoiding this error and more closely resembling our target use-case.  The assessments in this paper used GPS trajectories obtained from the GeoLife project~\cite{zheng2008understanding} which include ground truth (note that ground truth data for our target application was unavailable at the time of this writing). All experiments were performed on a 2000 MHz AMD EPYC 7713 CPU, and an NVIDIA GA100 GPU using Python 3.10 and PyTorch.  Source code is available via \url{https://github.com/lab-v2/Error-Detection-and-Correction}.
\vspace{3pt}

\begin{table}[t]
    \begin{tabular}{|l|l|l|l|}
    \hline
        Evaluated & Error Precision &  Error Recall & Error F1 \\
        Model & via EDCR &  via EDCR & via EDCR \\
        \hline
        LRANa & 0.999 & 0.941 & 0.969\\ \hline
        LRCN & 0.996 & 0.780 & 0.875\\ \hline
        CNN & 0.987 & 0.982 & 0.984\\ \hline
    \end{tabular}
     \caption{EDCR Error Detection Results - this table shows EDCR's ability to detect error for three different models.}
\label{tab:errorDetResultTable}
\end{table}
\begin{figure}[ht]
\centering
\includegraphics[width=1.07\columnwidth]{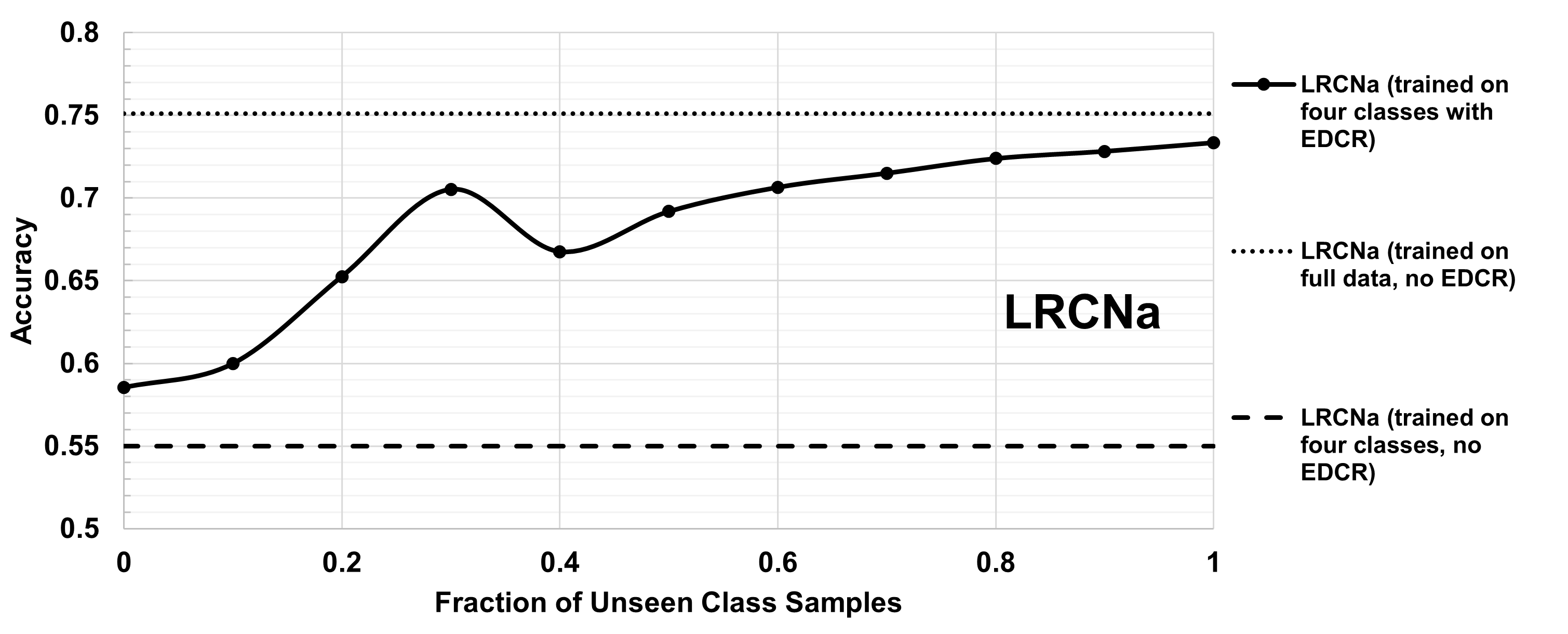}
\includegraphics[width=1.07\columnwidth]{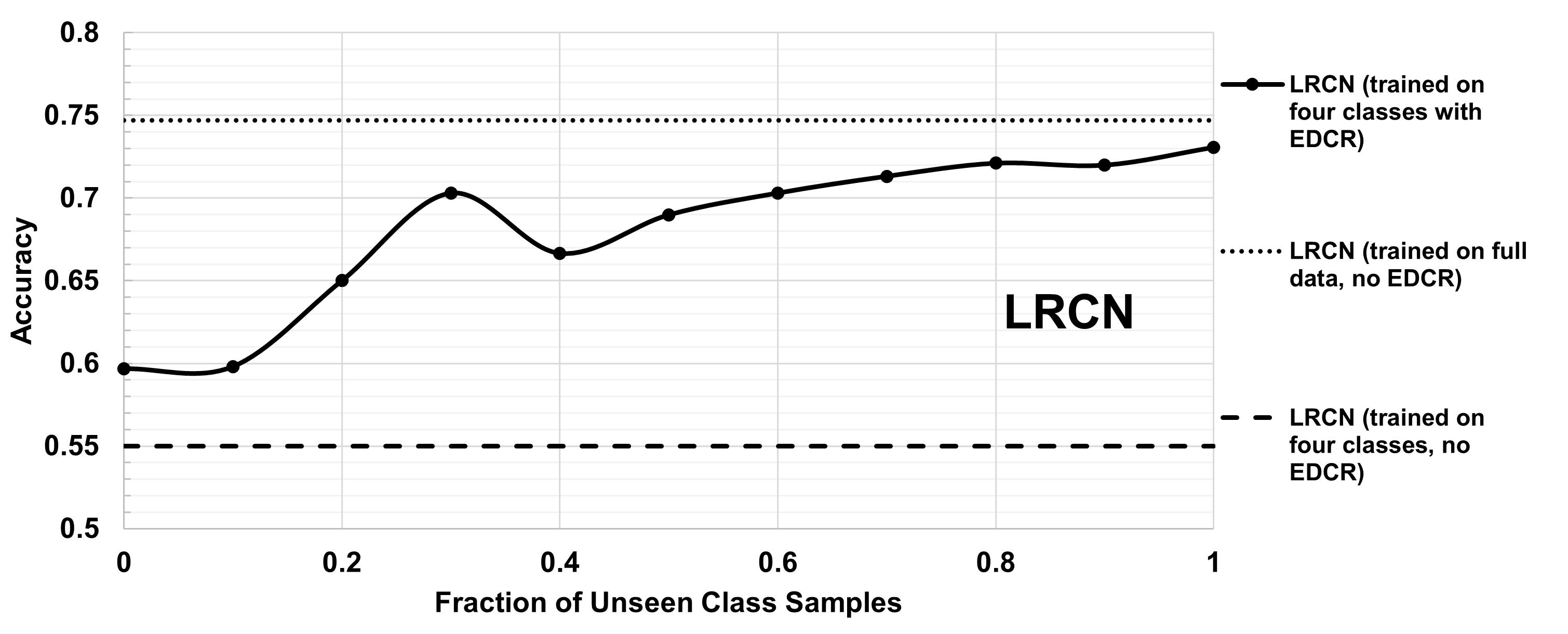}
\includegraphics[width=1.07\columnwidth]{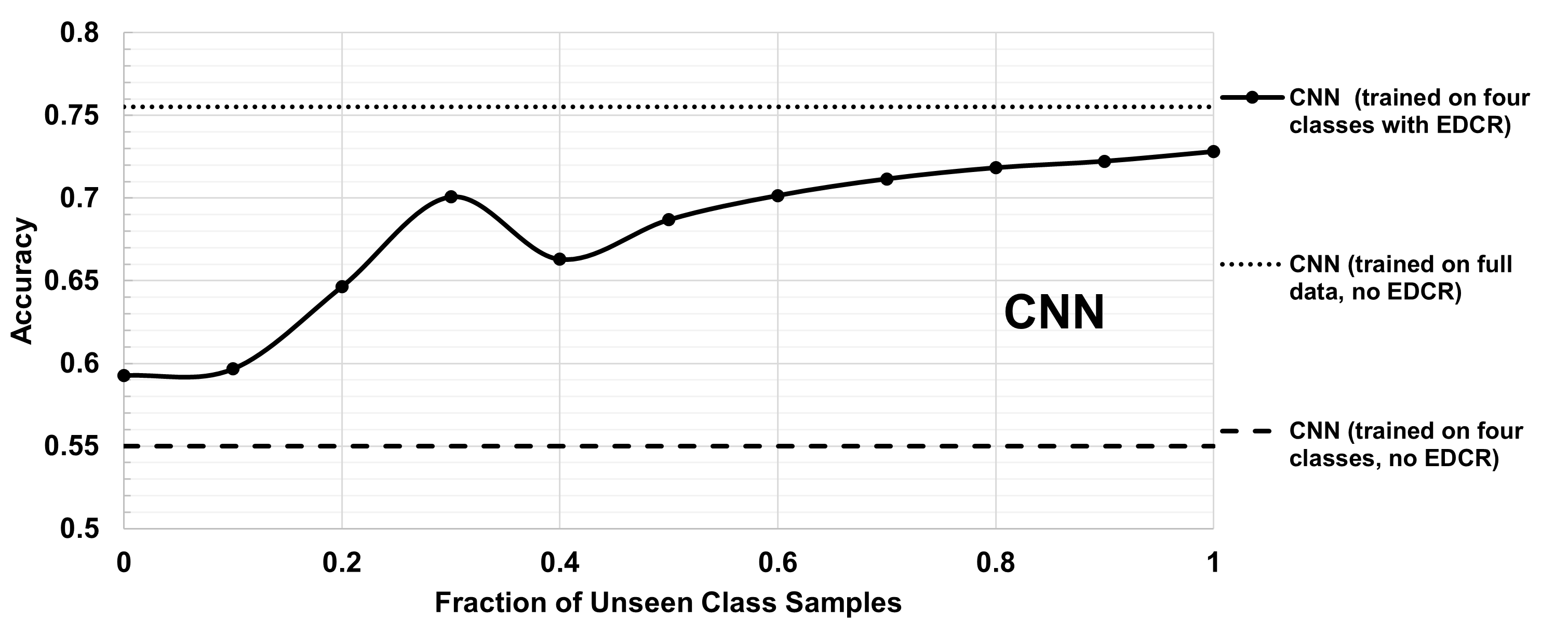}
\caption{Results for experiments with two movement classes removed from training for the LRCNa, LRCN, and CNN models.}
\label{fig:unseen}
\end{figure}

\begin{figure*}[h!]
\centering
\includegraphics[width=0.33\textwidth]{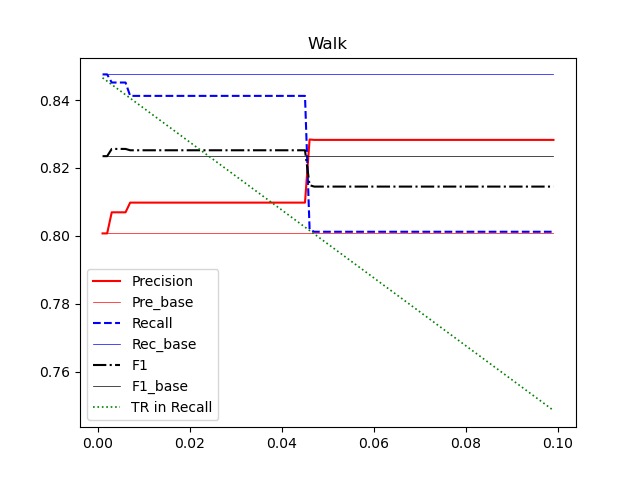}
\includegraphics[width=0.33\textwidth]{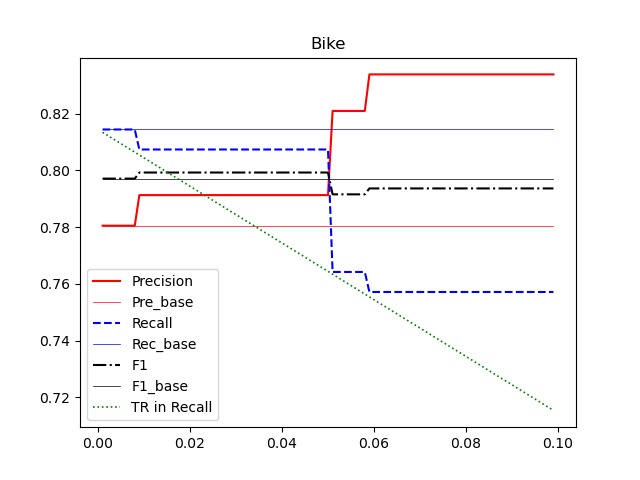}
\includegraphics[width=0.33\textwidth]{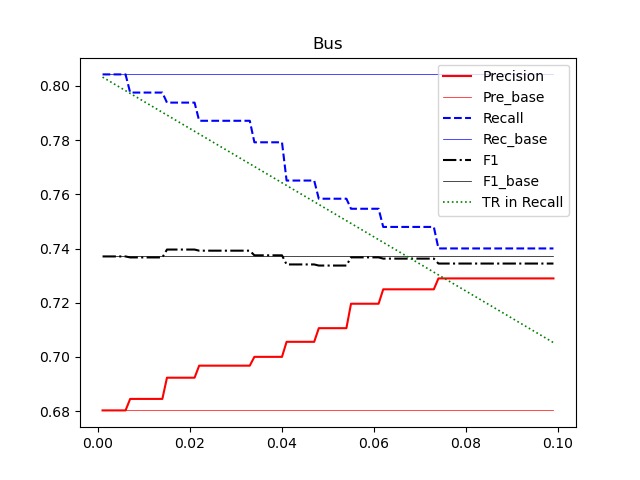}
\includegraphics[width=0.33\textwidth]{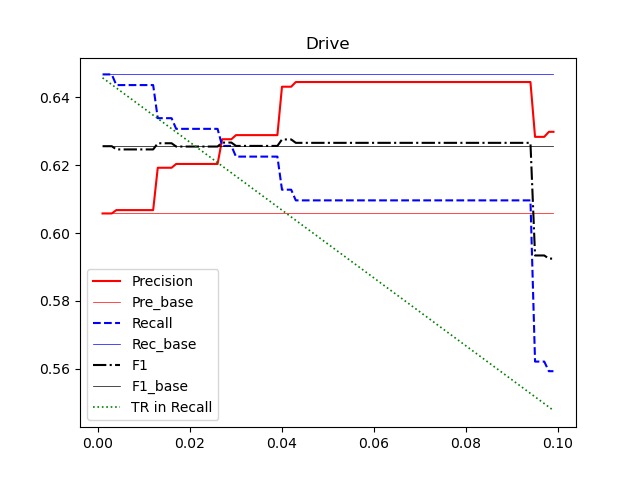}
\includegraphics[width=0.33\textwidth]{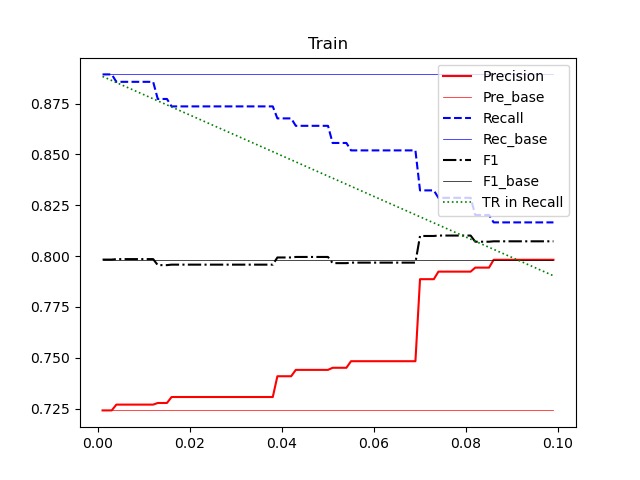}
\caption{LRCN Results for application of error detection and correction rules as a function of $\epsilon$.  TR in Recall is the theoretical reduction in recall based on analytic results.}
\label{fig:lrcnabyclass}
\end{figure*}

\noindent\textbf{Error Detection Experiments.}  First we examined the ability of learned error detection rules to detect errors in the underlying model.  Here we examined three base model architectures CNN~\cite{dabiri2018inferring}, LRCN~\cite{kim2022gps}, and our version of LRCN with an additional attention head (LRCNa).  In this experiment, error detection rules were trained from the same training data as the model.  Similar to previous work on examining the ability to detect errors in a machine learning model~\cite{daftry_introspective_2016} we evaluated precision, recall, and F1 of the ability of rules to identify errors.  These can be thought of as the fraction of results where our learned error detection rules correctly return an error (error precision), the fraction of errors identified (error recall), and the harmonic mean of the two (error F1).  The results shown in Table~\ref{tab:errorDetResultTable} demonstrate consistently high precision and recall for detecting errors across all model types - specifically obtaining a $0.875$ F1 for errors in the SOTA model (LRCN) and a top F1 of $0.984$ (for CNN).
\vspace{3pt}

\noindent\textbf{Test Data with Additional Classes.}  A key set of concerns for our use-case was the ability to deploy movement trajectory classification in an environment where the data differs from the training data - specifically containing previously unidentified classes.  To examine this, we trained CNN, LRCN, and LRCNa models without incorporating the \textsf{walk} and \textsf{drive} classes (Figure~\ref{fig:unseen}).  We note here both detection and correction are used. We initially learned the EDCR rules with the same training data in the model - which results in no sample being corrected to a class unseen in training data and effectively is zero-shot tuning of the base model by EDCR.  However, due to detection, this still resulted in accuracy improvements of $6.41\%$, $8.51\%$, and $7.76\%$ for LRCNa, LRCN, and CNN respectively.  We then added few-shot samples from the unseen data (the x-axis of Figure~\ref{fig:unseen}) giving us few-shot tuning of the base model.  Here with only $20\%$ of the samples with the unseen classes, we obtained an overall accuracy of $0.65$ on all three models representing a $17-18\%$ improvement.  We note these results are obtained without direct access to the underlying model, which may indicate that EDCR has the potential for adaptation of arbitrary $f_\theta$ models to novel scenarios - a key use case for our government customer.
\vspace{1.5pt}

\noindent\textbf{Precision-Recall Trade-off.}  A key intuition in our algorithmic design with the ability for the hyperparameter to $\epsilon$ to trade-off precision and recall.  Hence, we examined the effect in varying $\epsilon$ on test data that resembled training data (results for LRCN are shown in Figure~\ref{fig:lrcnabyclass}).  Recall that $\epsilon$ is interpreted as the maximum decrease in recall. We observed and validated the theoretical reduction (TR) in recall empirically and the experiments show us that in all cases, recall was no lower than the threshold specified by the hyperparameter $\epsilon$ though recall decreases as $\epsilon$ increases. In many cases, the experimental evaluation reduced recall significantly less than expected. We also see a clear relationship between $\epsilon$, precision, and recall: increasing $\epsilon$ leads to increased precision and decreased recall - which also aligns with our analytical results.  We also note that while \textsf{DetCorrRuleLearn} calls for a single $\epsilon$ hyperparameter, it is possible to set it differently for each class (e.g., lower values for classes where recall is important, higher values for classes where false positives are expensive). This may be beneficial as F1 for different classes seemed to peak for different values of $\epsilon$. We leave the study of heterogeneous $\epsilon$ settings to future work.
\vspace{1.5pt}

\begin{table}[h!]
\vskip 0.15in
\begin{center}
    \begin{tabular}{|l|l|l|}
    \hline
        Evaluated & No EDCR & With EDCR \\
        Model & (baseline) & (ours)\\
        \hline
        LRCNa  &  0.751& \textbf{0.763} (+1.6\%)\\ \hline
        LRCN &  0.747 &\textbf{0.760} (+1.7\%)  \\ \hline
        CNN & 0.755 &0.755 ($\pm$ 0\%) \\ \hline
    \end{tabular}
\caption{Overall accuracy when all classes are represented with and without EDCR.}  
\label{tab:accuracyRegular}
\end{center}
\end{table}

\noindent\textbf{Accuracy Improvement via EDCR.}  We also investigated EDCR's ability to provide overall accuracy improvement to the base model.  Here we trained each of the three models (LRCNa, LRCN, CNN) and associated EDCR rules (on the same training data as the model) and evaluated the overall accuracy on the test set both with and without applying rules (see Table~\ref{tab:accuracyRegular}).  We found that that EDCR provided a noticeable improvement in both LRCN and LRCNa models - effectively establishing a new SOTA when evaluated with no overlap between training and testing. We also examined other splits between training and testing (not depicted) and obtained comparable results.

\section{Conclusion} 
We propose a rule-based framework for the error detection and correction of supervised neural models for classification of movement trajectories. Our framework uses the training data to learn rules to be employed in the testing phase. Firstly, we use the detection rules to identify the movement trajectories that are misclassified by the supervised model and then we use the correction rules to re-classify the movement. Further, we formally prove the relation of confidence and support of the learned rules to the changes in the classification metrics like precision and recall. To show EDCR's emperical validation, we first report the framework's ability to identify errors with the F1 scores going up to $0.984$. We also show overall accuracy imporvement over the SOTA model by employing the EDCR framework. Our framework is specifically useful in cases of encountering novel classes not seen in training data as shown by a $8.51\%$ improvement of unseen movement accuracy over SOTA for zero-shot tuning. Additionally, we discuss operationalizing our trajectory classification method in our deployed system.
There are several directions for future work.   First, we look to explore other methods to create the conditions, in particular leveraging ideas from conformal prediction~\cite{NEURIPS2023_fe318a2b}. Another direction is to look at alternative solutions to learn the rules allowing for more complicated rule structures.  Finally, the use of rules for error detection and correction of machine learning models presented here may be useful in domains such as vision. To reliably incorporate vision models in real-world applications for tasks like object detection, image classification, and motion tracking, etc., EDCR framework can be leveraged to improve the overall system's accuracy and robustness by identifying and correcting it's misclassification.

\section*{Ethical Statement}

There are no ethical issues.

\section{Acknowledgments}
This research is supported by the Intelligence Advanced Research Projects Activity (IARPA) via the Department of Interior/ Interior Business Center (DOI/IBC) contract number 140D0423C0032. The U.S. Government is authorized to reproduce and distribute reprints for Governmental purposes notwithstanding any copyright annotation thereon. Disclaimer: The views and conclusions contained herein are those of the authors and should not be interpreted as necessarily representing the official policies or endorsements, either expressed or implied, of IARPA, DOI/IBC, or the U.S. Government.  Additionally, some of the authors are supported by ONR grant N00014-23-1-2580 and ARO grant  W911NF-24-1-0007.

\bibliographystyle{named}
\bibliography{main}

\end{document}